\documentclass[11pt,a4paper]{article}
\usepackage[hyperref]{naaclhlt2018}
\usepackage{times}
\usepackage{latexsym}

\usepackage{url}
\usepackage{graphicx}

\usepackage{tikz}
\usetikzlibrary{trees}
\usepackage{verbatim}
\usepackage{pgfplots}
\usepackage{adjustbox}
\usepackage{amsmath}

\usepackage{hyperref}

\aclfinalcopy 

\title{ETH-DS3Lab at SemEval-2018 Task 7:  Effectively Combining Recurrent and Convolutional Neural Networks for Relation Classification and Extraction}

\author{Jonathan Rotsztejn\textsuperscript{1}, Nora Hollenstein\textsuperscript{1,2}, Ce Zhang\textsuperscript{1} \\
  \textsuperscript{1} Systems Group, ETH Zurich \\
  {\tt \{rotsztej,noraho\}@ethz.ch, ce.zhang@inf.ethz.ch}\\ 
  \textsuperscript{2} IBM Research, Zurich
  }

\date{}

\begin{document}
\maketitle
\begin{abstract}
  Reliably detecting relevant relations between entities in unstructured text is a valuable resource for knowledge extraction, which is why it has awaken significant interest in the field of Natural Language Processing. 
  In this paper, we present a system for relation classification and extraction based on an ensemble of convolutional and recurrent neural networks that ranked first in 3 out of the 4 Subtasks at SemEval 2018 Task 7. We provide detailed explanations and grounds for the design choices behind the most relevant features and analyze their importance.
\end{abstract}

\section{Introduction and related work}\label{intro}

One of the current challenges in analyzing unstructured data is to extract valuable knowledge by detecting the relevant entities and relations between them. The focus of SemEval 2018 Task 7 is on relation classification (assigning a type of relation to an entity pair - \textit{Subtask 1}) and relation extraction (detecting the existence of a relation between two entities and determining its type - \textit{Subtask 2}). 

\begin{figure}[t!]
    \centering
    \includegraphics[width=0.45\textwidth]{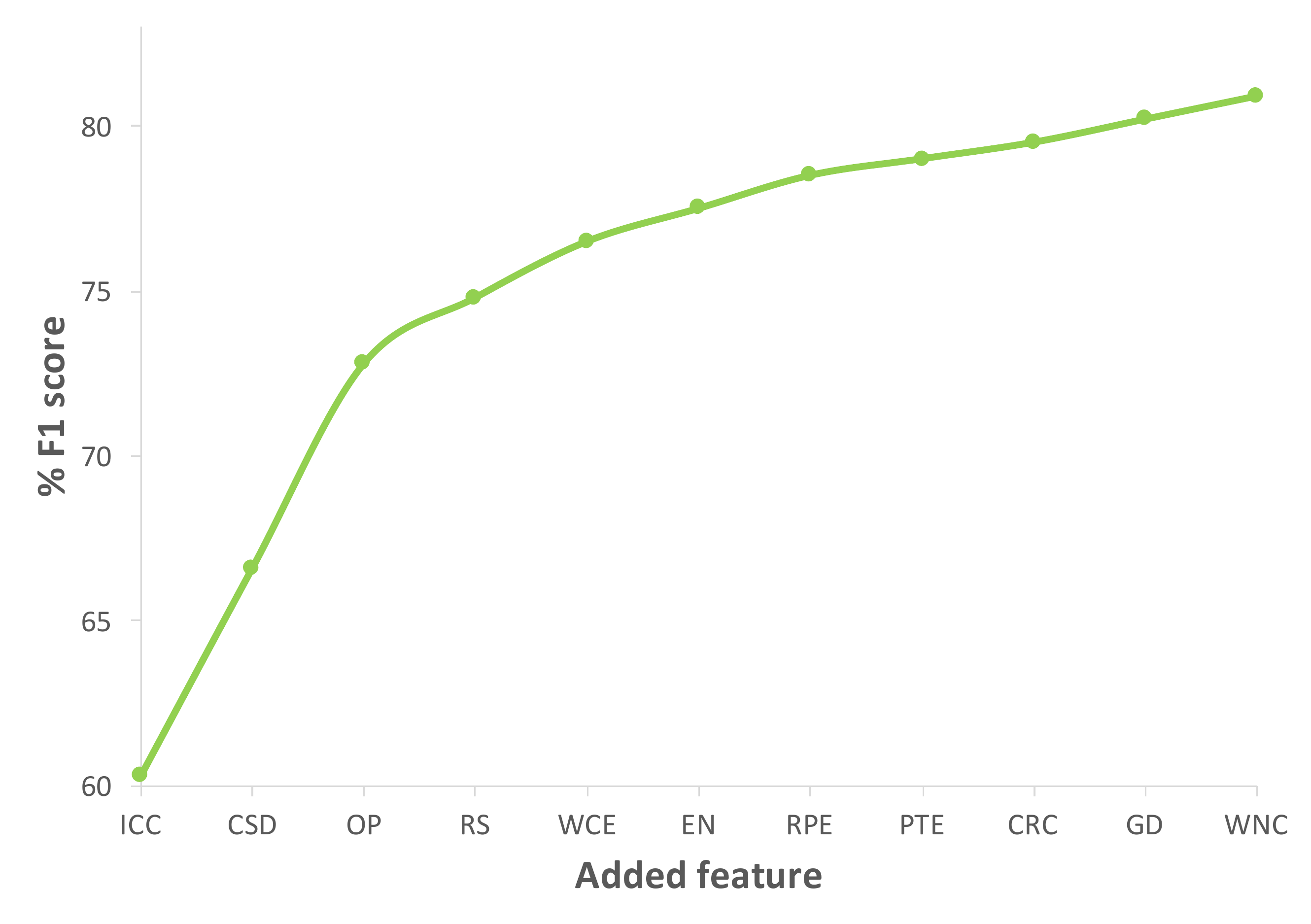}
    \includegraphics[width=0.5\textwidth]{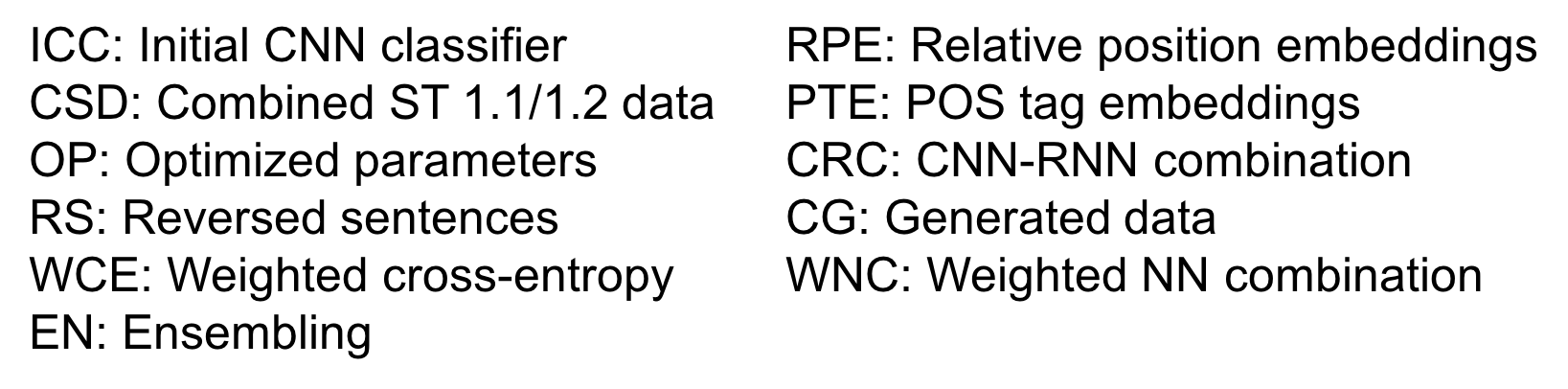}
    \caption{Feature addition study to evaluate the impact of the most relevant features on the $F_1$ score of the 5-fold cross-validated training set of Subtasks 1.1 and 1.2}
    \label{feature_addition}
\end{figure}

Moreover, the task distinguishes between relation classification on clean data (i.e.: manually annotated entities - \textit{Subtask 1.1}) and noisy data (automatically annotated entities - \textit{Subtask 1.2}). It addresses semantic relations from 6 categories, all of them specific to scientific literature. Relation instances are to be classified into one of the following classes: USAGE, RESULT, MODEL-FEATURE, PART-WHOLE, TOPIC, COMPARE, where the first five are asymmetrical relations and the last is order-independent (see \citet{taskpaper} for a more detailed description of the task).
Since the training data was provided by the task organizers, we focused on supervised methods for relation classification and extraction. Similar systems in the past have been based on Support Vector Machines \cite{uzuner20112010,minard2011multi}, Na\"ive Bayes \cite{zayaraz2015concept} and Conditional Random Fields \cite{sutton2006introduction}. More recent approaches have experimented with neural network architectures \cite{socher2012semantic,fu2017domain}, especially convolutional neural networks (CNNs) \cite{nguyen2015relation,lee2017semeval} and recurrent neural networks (RNNs) based on LSTMs \cite{zheng2017joint, peng2017cross}. The system presented in this article builds upon the latest improvements in employing neural networks for relation classification and extraction. An overview of the most relevant features is shown on Figure \ref{feature_addition}.

\section{Method}

\subsection{Neural architecture}

\begin{figure*}[t!]
    \centering
    \includegraphics[width=\textwidth]{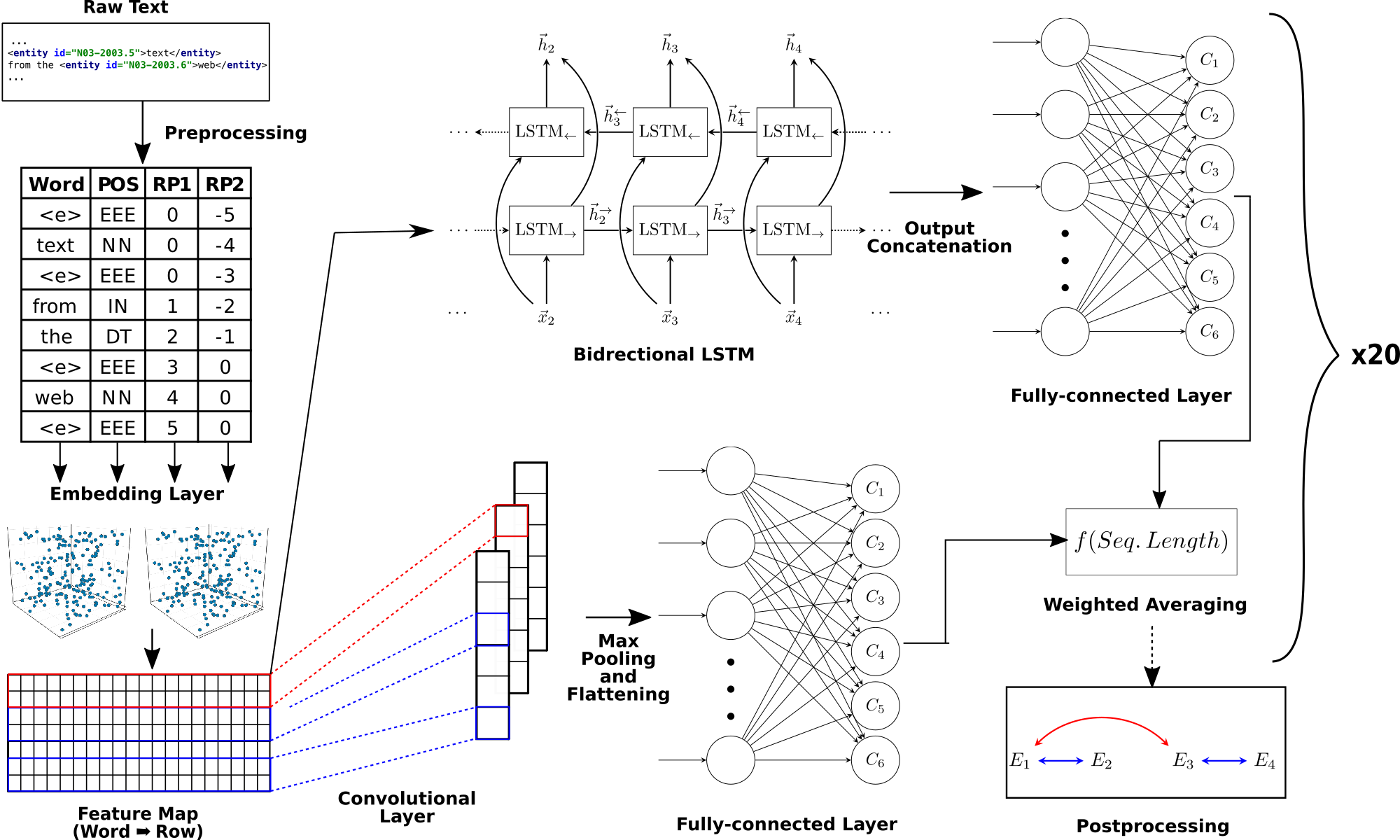}
    \caption{Full pipeline architecture}
    \label{architecture}
\end{figure*}
Figure \ref{architecture} shows the full architecture of our system. Its main component is an ensemble of CNNs and RNNs. 
The CNN architecture follows closely on \cite{kim2014convolutional,collobert2011natural}. It consists of an initial embedding layer, which is followed by a convolutional layer with multiple filter widths and feature maps with a ReLU activation function, a max-pooling layer (applied over time) and a fully-connected layer, that is trained with dropout, and produces the output as logits, to which a softmax function is applied to obtain probabilities.
The RNN consists of the same initial embedding layer, followed two LSTM-based sequence models \cite{hochreiter1997long}, one in the forward and one in the backward direction of the sequence, which are dynamic (i.e.: work seamlessly for varying sequence lengths). The output and final hidden states of the forward and backward networks are then concatenated to a single vector. Finally, a fully-connected layer, trained with dropout, connects this vector to the logit outputs, to which a softmax function is applied analogously to obtain probabilities.

The complete architecture was replicated and trained independently several times (see Table \ref{params}) using different random seeds that ensured distinct initial values, sample ordering, etc. in order to form an ensemble of classifiers, whose output probabilities were averaged to obtain the final probabilities for each class. We analyzed and tried several deeper and more complex neural architectures, such as multiple stacked LSTMs (up to 4) and models with 2 to 4 hidden layers, but they didn't achieve any significant improvements over the simpler models. Conclusively, the strategy that produced the best results consisted of  adequately combining the individual predictions of the single models (see section \ref{combining}). 

\subsection{Domain-specific word embeddings}\label{nlp_emb}
\begin{figure}[t!]
    \centering
    \includegraphics[width=0.45\textwidth]{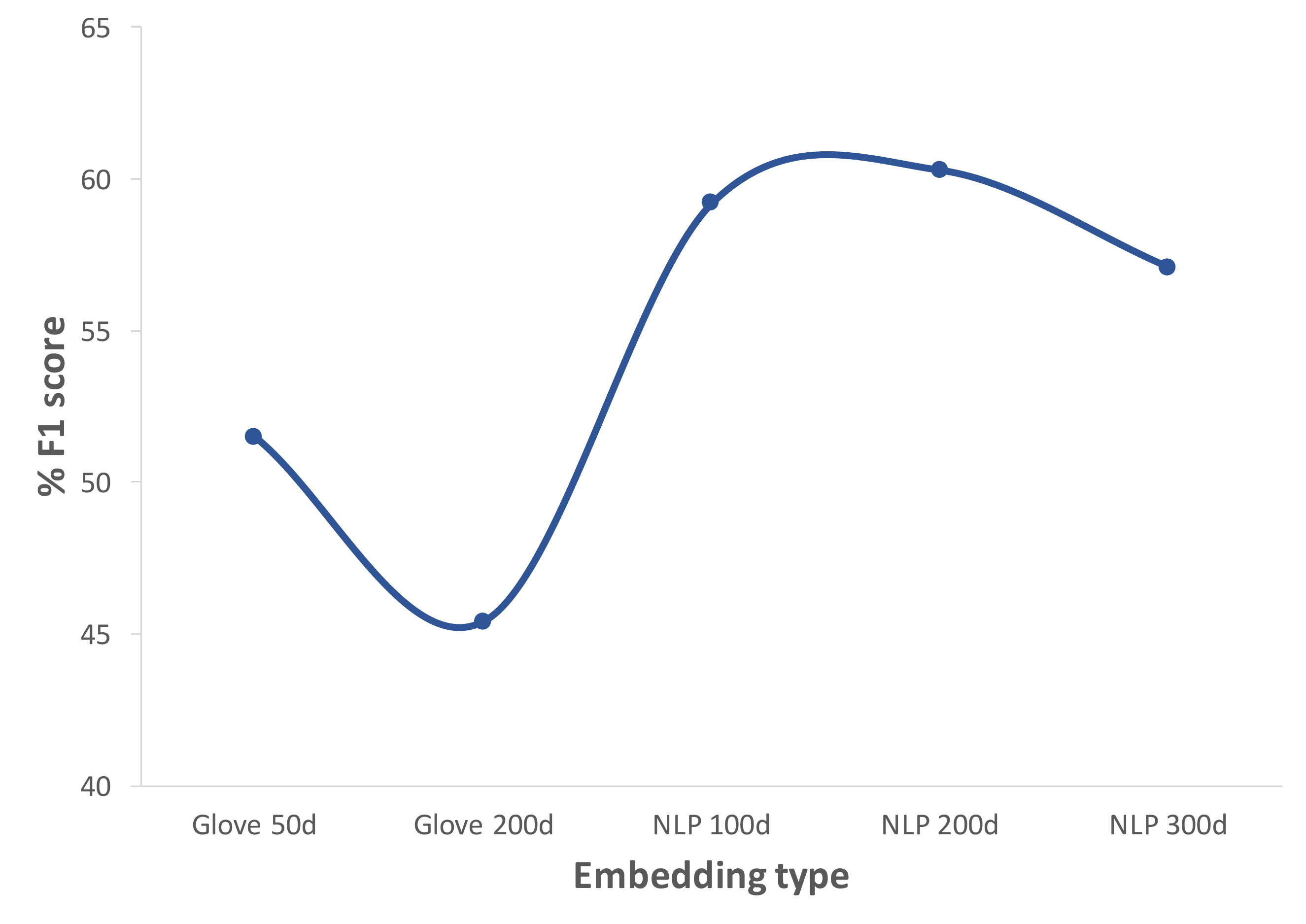}
    \caption{Effect of different word embedding types based on a simple CNN classifier for Subtask 1.1}
    \label{embeddings}
\end{figure}

We collected additional domain-specific data from scientific NLP papers to train word embeddings. All \textit{ArXiv cs.CL} abstracts since 2010 (1 million tokens) and the \textit{ACL ARC corpus} (90 million tokens; \citet{bird2008acl}) were downloaded and preprocessed. We used \textit{gensim} \cite{rehurek_lrec} to train word2vec embeddings on these two data sources, and additionally the sentences provided as training data for the SemEval task (in total: 91,304,581 tokens). We experimented with embeddings of 100, 200 and 300 dimensions, where 200 dimensions yielded the best performance for the task as shown in Figure \ref{embeddings}.

\subsection{Preprocessing}\label{preprocessing}
\begin{figure}[t!]
    \centering
    \includegraphics[width=0.45\textwidth]{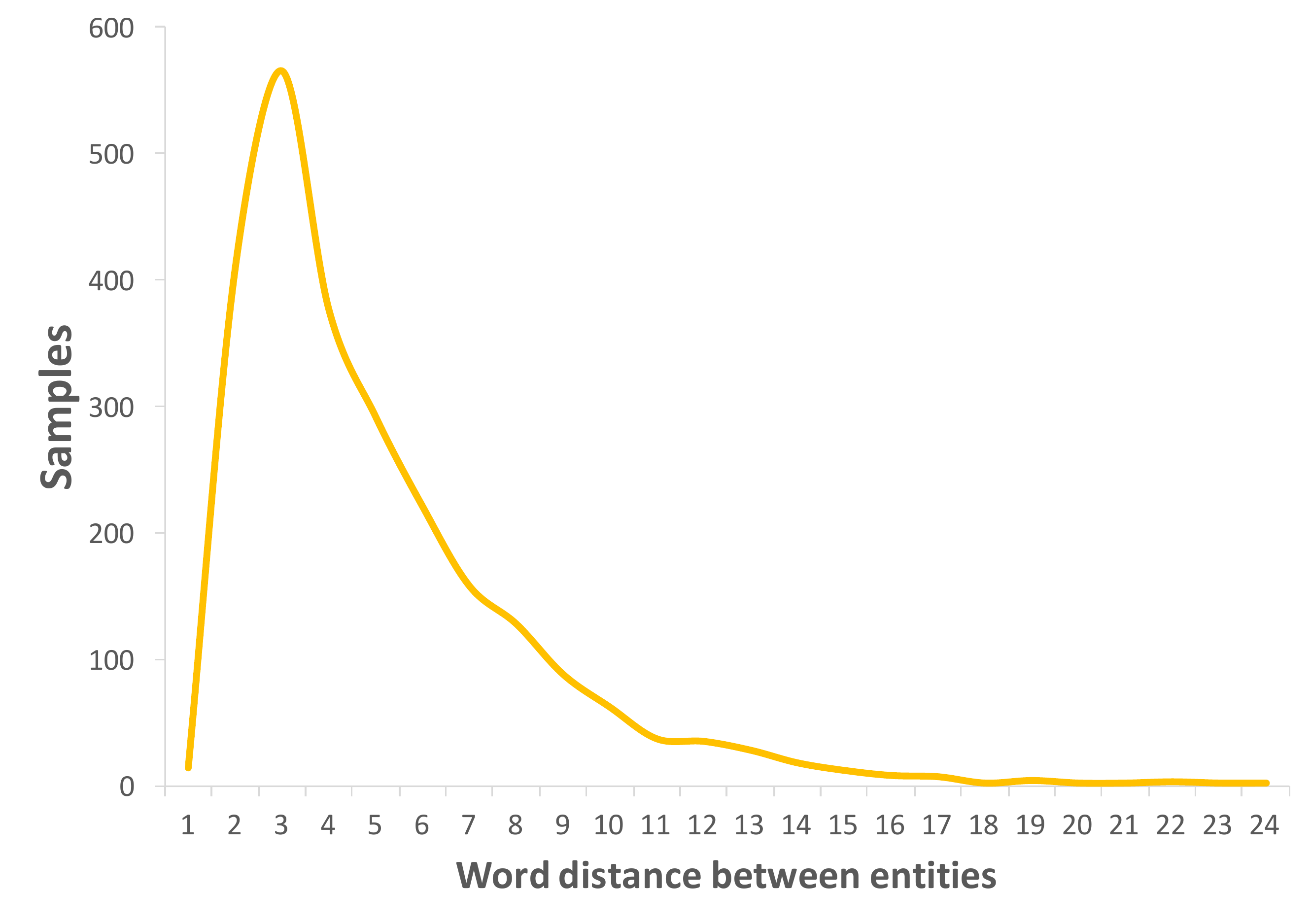}
    \caption{Word distance between entities in a relation for training data in Subtask 1.1}
    \label{dist}
\end{figure}

\paragraph{Cropping sentences}
Since the most relevant portion of text to determine the relation type is generally the one contained between and including the entities \cite{lee2017semeval}, we solely analyzed that part of the sentences and disregarded the surrounding words.
For Subtask 2, we initially considered every entity pair contained within a single sentence as having a potential relation. Since the probability that a relation between two entities exists drops very rapidly with increasing word distance between them (see Figure \ref{dist}), we only considered sentences that didn't exceed a maximum length threshold (see Table \ref{params}) between entities to diminish the chances of predicting false positives in long sentences. 

\begin{figure}[t!]
    \centering
    \includegraphics[width=0.45\textwidth]{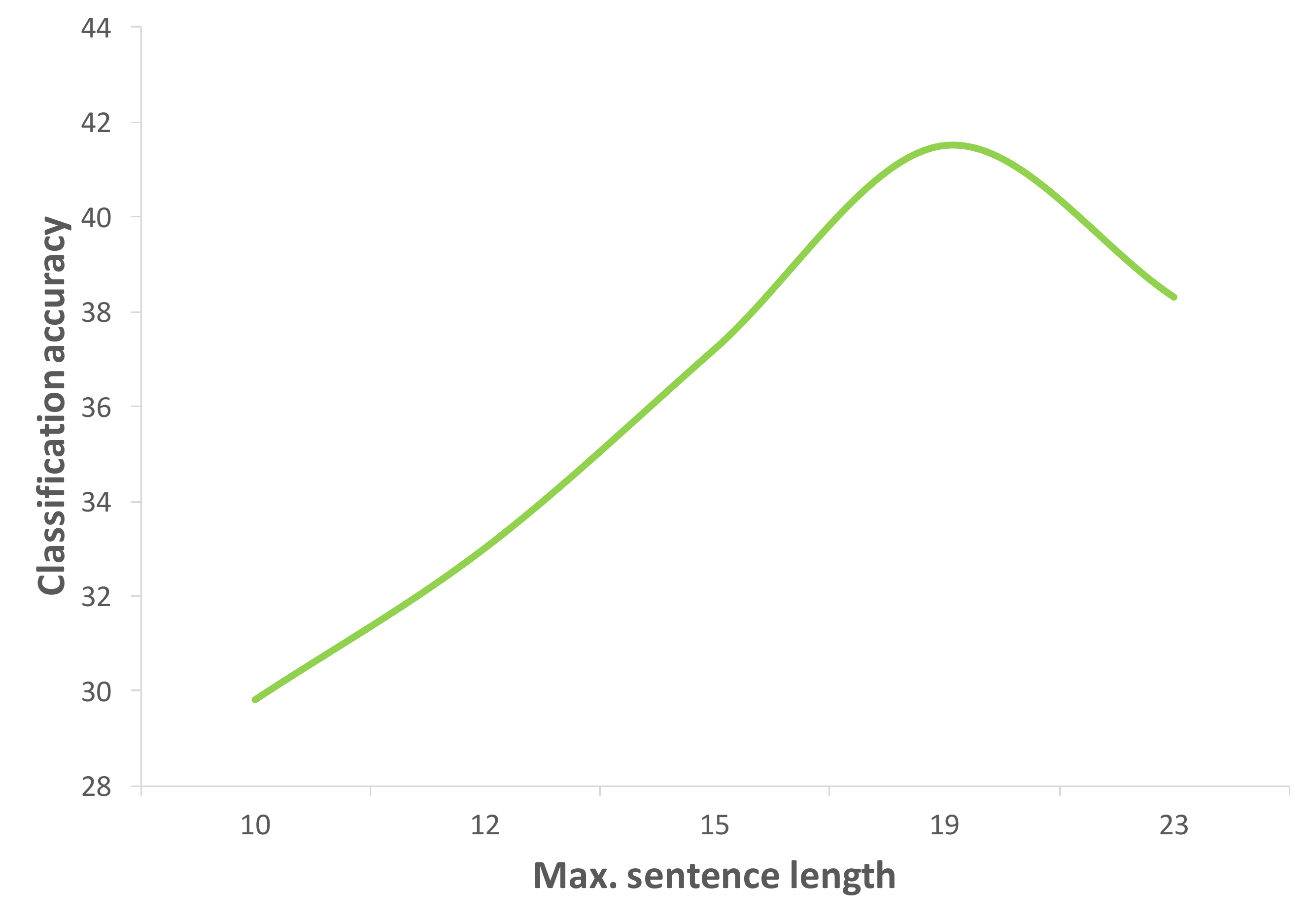}
    \caption{Effect of max. length threshold on accuracy for a preliminary RNN-based classifier}
    \label{max_sentence_length}
\end{figure}

Various experiments with different thresholds between 7 and 23 words on the training set showed that the best results on sentences from scientific papers are achieved with a threshold of 19 words, as shown in Figure \ref{max_sentence_length}.

\paragraph{Cleaning sentences} 
Some of the automatically annotated samples contained nested entities such as \textit{\small{\textless entity id="L08-1220.16"\textgreater ~signal \textless entity id="L08-1220.17"\textgreater ~processing \textless /entity\textgreater \textless /entity\textgreater}}. We flattened these structures into simple entities and considered all the entities separately for each train and test instance.  Moreover, all tokens between brackets [] and parentheses () were deleted, and the numbers that were not part of a proper noun replaced with a single wildcard token.
\looseness=-1

\paragraph{Using entity tags} 
In order to provide the neural networks with explicit cues of where an entity started and ended, we used a single symbol, represented as an XML tag  \texttt{<e>} before and after the entity, to indicate it \cite{dligach2017neural}. 

\paragraph{Relative order strategy \& number of classes}\label{relative}
As mentioned in Section \ref{intro}, 5 out of the 6 relation types are asymmetrical and the tagging is always done by using the same order for the entities as the one found in the abstracts' text/title. For that reason, it was important to carefully devise a schema that allowed generalization by exploiting the information from both ordered and reversed (words that will be treated here as antonyms) relations. Apart from using the relative position embeddings presented by \citet{lee2017semeval}, for Subtask 1, we incorporated a full text reversal of those sentences in which a reverse relation was present, both at training and testing time. The result were instances that, although not corresponding to a valid English grammar, frequently resembled more in structure to their ordered counterparts. This has been illustrated by an example of two instances belonging to the \textit{PART-WHOLE} class in Figure \ref{reversed_sentence}. 

\tikzstyle{level 1}=[level distance=1.5cm, sibling distance=3.5cm]
\tikzstyle{level 2}=[level distance=1.2cm, sibling distance=3.5cm]

\tikzstyle{bag} = [text width=20em, text centered]

\begin{figure}[t!]
\begin{center}
\begin{tikzpicture}[grow=down, edge from parent/.style={draw,-latex}, font={\fontsize{9pt}{12}\selectfont}]
\node[bag] {\textless e\textgreater ~corpus \textless  e\textgreater ~consists of independent \textless e\textgreater  ~text \textless e\textgreater}
    child {
        node[bag] {\textless e\textgreater ~text \textless e\textgreater ~independent of consists \textless e\textgreater ~corpus \textless e\textgreater}
            child {
                    node[bag] {\textless e\textgreater ~texts \textless e\textgreater ~from a \textless e\textgreater ~target corpus \textless e\textgreater}
                    edge from parent[<->, dashed]
                    node[right] {$Resembles$}
            }
            edge from parent 
            node[right] {$REVERSE$}
    }
;
\end{tikzpicture}
\end{center}
\caption{Example of a reversed sentence}
\label{reversed_sentence}
\end{figure}
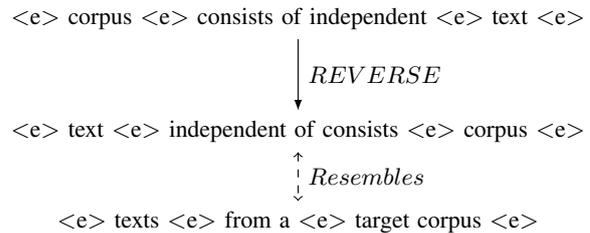

Thus, the system could operate by using only the 6 originally specified relation types and merely learn how to identify ordered relations, rather than having to handle the two different types of patterns or to add extra classes to describe both the ordered and the reversed versions of each class, which helped improve the overall accuracy of the classifier (+2.0\% $F_1$).

For Subtask 2, since no information regarding the ordering of the arguments was available (the extraction and the ordering were part of the task), we opted for a 12-class strategy: one for each of the 5 ordered and reversed relations, plus the symmetrical relation (COMPARE) and a NONE class for the negative instances, i.e.: those that didn't contain any relation at all. An alternative 6-class approach based on presenting the sentences both ordered and reversed to the network, computing two predictions for each and afterwards consolidating both did not produce good results (-3.4\% $F_1$).
\looseness=-1

\paragraph{Part-of-speech tags}
We used the Stanford CoreNLP tagger \cite{manning-EtAl:2014:P14-5} to obtain POS tags for each word in every sentence in the dataset and trained high-dimensional embeddings for the 36 possible tags defined by the Penn Treebank Project \cite{marcus1993building}. Moreover, the XML tags to identify the entities and the number wildcard received their own corresponding artificial POS tag embedding (see Figure \ref{architecture} for a detailed example).

\section{Experiments}

\subsection{Exploiting provided data}
One of the main challenges of the task was the limited size of the training set, which is a common drawback for many supervised novel machine learning tasks. To overcome it, we combined the provided datasets\footnote{\href{https://groups.google.com/forum/\#!topic/semeval18task7/XrQbsIWvxpE}{Link to forum post 1} -  \href{https://groups.google.com/forum/\#!topic/semeval18task7/72ye1uH9QKI}{Link to forum post 2}} for Subtask 1.1 and 1.2 to train the models for both Subtasks (+6.2\% $F_1$). Furthermore, we leveraged the predictions of our system for Subtasks 1.1 and 1.2 and added them as training data for Subtask 2 (+3.6\% $F_1$).

\subsection{Generating additional data}
Due to the limited number of training sentences provided, we explored the following approach to augment the data: We generated automatically-tagged artificial training samples for Subtask 1 by combining the entities that appeared in the test data with the text between entities and relation labels of those from the training set (see Table \ref{sample}). To evaluate the quality of the sentences and augment our data only with sensible instances, we estimated an NLP language model using the KenLM Language Model Toolkit \cite{heafield2011kenlm} on the corpus of NLP-related text described in Section \ref{nlp_emb} and evaluated the generated sentences with it. Furthermore, we set a minimum threshold of 5 words for the length of the text between entities, limited the number of sentences generated from each of them to a single instance in order to promote variety, and only kept those sentences that score a very high probability (-21 in log scale) against the language model. This process yielded 61 additional samples on the development set (+0.7\% $F_1$).

\begin{table*}[]
\centering
\begin{tabular}{|l|l|}
\hline
Dev set:    & \textit{\textless e\textgreater~predictive performance \textless e\textgreater~of our \textless e\textgreater~models \textless e\textgreater}  \\\hline
Train set:  & \textit{\textless e\textgreater~methods \textless e\textgreater~involve the use of probabilistic \textless e\textgreater~generative models \textless e\textgreater}    \\\hline\hline
New sample: & \textit{\textless e\textgreater~predictive performance \textless e\textgreater~involve the use of probabilistic \textless e\textgreater~models \textless e\textgreater}\\\hline
\end{tabular}
\caption{Generated sample}
\label{sample}
\end{table*}

\subsection{Parameter optimization}\label{opt}
To determine the optimal tuning for our richly parameterized models, we ran a grid search over the parameter space for those parameters that were part of our automatic pipeline. The final values and evaluated ranges are specified in Table \ref{params}.

\begin{table*}[h]
\centering
\begin{tabular}{|l|l|l|}
\hline
\textbf{Parameter}                                        & \textbf{Final value} & \textbf{Experiment range} \\\hline\hline
Word embedding dimensionality                    & 200         & 100-300          \\
Embedding dimensionality for part-of-speech tags & 30          & 10-50            \\
Embedding dimensionality for relative positions  & 20          & 10-50            \\
Number of CNN filters                            & 192         & 64-384           \\
Sizes of CNN filters                             & 2 to 7      & 2-4 to 5-9       \\
Norm regularization parameter ($\lambda$)           & 0.01        & 0.0-1.0          \\
Number of LSTM units (RNN)                       & 600         & 0-2400           \\
Dropout probability (CNN and RNN)                & 0.5         & 0.0-0.7          \\
Initial learning rate              & 0.01        & 0.001-0.1        \\
Number of epochs (Subtask 1)                     & 200         & 20-400             \\
Number of epochs (Subtask 2)                     & 10         & 5-40              \\
Ensemble size                                    & 20          & 1-30             \\
Training batch size                              & 64          & 32-192           \\
Upsampling ratio (only Subtask 2)                & 1.0           & 0.0-5.0           \\
Max. sentence length (only subtask 2) & 19  & 7-23 \\\hline  
\end{tabular}
\caption{Final parameter values and their explored ranges}
\label{params}
\end{table*}

\subsection{Defining the objective}\label{defining}
The cross-entropy loss, defined as the cross-entropy between the probability distribution outputted by the classifier and the one implied by the correct prediction is one of the most widely used objectives for training neural networks for classification problems \cite{janocha2017loss}. A shortcoming of this approach is that the cross-entropy loss usually only constitutes a conveniently decomposable proxy for what the ultimate goal of the optimization is \cite{eban2017scalable}: in this case, the macro-averaged $F_1$ score. Motivated by the fact that individual instances of infrequent classes have a bigger impact on the final $F_1$ score than those of more frequent ones \cite{manning2008introduction}, we opted for a weighted version of the cross-entropy as loss function, where each class had a weight \textit{w} that was inversely proportional to their frequency in the training set: 

\begin{equation*}
    w_{class\ i} = \frac{\sum_j\#_{class\ j}}{N_{classes}*\#_{class\ i}}
\end{equation*}

\noindent where $\#$ indicates the count for a certain class and $N_{classes}$ is the total number of classes.\\
The weights are scaled as to preserve the expected value of the factor $k_i$ that accompanies the logarithm in the mathematical expression of the loss formula: $L=-\sum k_i log(y_i)$, which is equal to $wy_{i}'$ for the weighted cross-entropy and $y_{i}'$ for the unweighted version, where $y_{i}'=1$ for the correct class and $y_{i}$ is the predicted probability for that class.
Illustrating this concept, it can be observed that a single instance of class \textit{TOPIC} (support of only 6 instances) could account for up to 2.8\% of the final score on the test set.
This function proved to be a better surrogate for the global final score than the standard cross-entropy (+1.6\% $F_1$). 

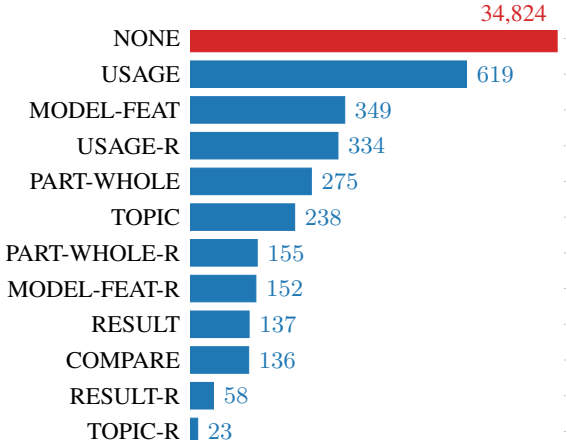
\begin{figure}[t!]
\setlength{\abovecaptionskip}{2pt}
\begin{tikzpicture}
  \tikzstyle{every node}=[font=\small]
  \definecolor{myred}{rgb}{0.839,0.153,0.157}
  \definecolor{myblue}{rgb}{0.122,0.467,0.706}
  \begin{axis}[
    xbar,
    height=7cm,
    y axis line style = { opacity = 0 },
    axis x line       = none,
    tickwidth         = 0pt,
    enlarge y limits  = 0.02,
    enlarge x limits  = 0.02,
    symbolic y coords = {TOPIC-R, RESULT-R, COMPARE, RESULT, MODEL-FEAT-R, PART-WHOLE-R, TOPIC, PART-WHOLE, USAGE-R, MODEL-FEAT, USAGE, NONE},
ytick={TOPIC-R, RESULT-R, COMPARE, RESULT, MODEL-FEAT-R, PART-WHOLE-R, TOPIC, PART-WHOLE, USAGE-R, MODEL-FEAT, USAGE, NONE},
    every axis plot/.append style={
          bar shift=0pt,
          fill
        },
    width=6.5cm
  ]
  
  \addplot[myblue, nodes near coords] coordinates { (619,USAGE) (349,MODEL-FEAT) (334,USAGE-R) (275,PART-WHOLE) (238,TOPIC) (155,PART-WHOLE-R) (152,MODEL-FEAT-R) (137,RESULT) (136,COMPARE) (58,RESULT-R) (23,TOPIC-R)};
  \addplot[myred, nodes near coords={34,824},  every node near coord/.append style={xshift=0pt,yshift=9pt,anchor=east}] coordinates { (820,NONE) };
  \end{axis}
\end{tikzpicture}
\caption{Class frequencies for Subtask 2}
\label{frequencies_st2}
\end{figure}

\subsection{Upsampling}
One of the challenges of our approach for Subtask 2 was the existence of a large imbalance between the target classes. Namely, the NONE class constituted the clear majority (Figure \ref{frequencies_st2}). To overcome it, we resorted to an upsampling scheme for which we defined an arbitrary ratio of positive to negative examples to present to the networks for the combination of all positive classes (+12.2\% $F_1$).
\looseness=-1

\section{Training and validating the model}
The neural networks were trained using an Adam optimizer with parameter values $\beta_2=0.9$, $\beta_2=0.999$, $\epsilon=1e-08$ (suggested default values in the TensorFlow library \cite{tensorflow2015-whitepaper}) with a step learning rate decay scheme on top of it. This consisted in halving the learning rate every 25 and 1 iterations through the whole dataset for Subtasks 1 and 2 respectively (note: the size of the upsampled dataset for Subtask 2 was about 25 times that of Subtask 1), starting from the initial value determined in Section \ref{opt}. In order to avoid overfitting the development set of each Subtask, we evaluated the quality of our models by applying a 5-fold cross-validation on the combined training data of Subtasks 1.1 and 1.2 and on the training data of Subtask 2.

\paragraph{Combining predictions}\label{combining}
During the development, we observed that similar $F_1$ scores could be achieved by using either a convolutional neural network or a recurrent one separately, but the combination of both outperformed the individual models. Moreover, since the RNN-based architecture had a tendency to obtain better results than its CNN-based counterpart for long sequences, we combined both predictions in such a way that a higher weight was assigned to the RNN predictions for longer sentences by applying: \hfill \vspace{3mm}
$w_{rnn,i} = 0.5 + sign(s_i)\cdot s_i^2$, where
\begin{equation*}
    s_i = \frac{length_i - \min_{j}(length_j)}{\max_{j}(length_j) - \min_{j}(length_j)}-0.5
\end{equation*}  and $length_i$ is the length of the i-th sentence.

\paragraph{Post-processing}
To enforce consistency with the text annotation scheme, some rules that were not built into the system had to be applied ex-post. First, predictions of reversed relations should not be of type COMPARE, since it is the only symmetrical relation. When this condition occurred, we simply predicted the class that had the 2nd highest probability. Second, each entity could only be part of one relation. To address this for Subtask 2, we run a conflict-solving algorithm that, in case of overlaps, always preferred short relations (cf. Figure \ref{embeddings}]) and broke ties by choosing the relation with the most frequent class in the training data and at random when it persisted.

\section{Results}

\subsection{Feature analysis}

We conducted a feature addition study to evaluate the impact of the most relevant features on the $F_1$ score of the 5-fold cross-validated training/development set of Subtasks 1.1 and 1.2. 

The results have been previously shown in Figure \ref{feature_addition}. It can be observed from the plot that substantial gains can be obtained by applying standalone data manipulation techniques that are independent of the type of classifier used, such as combining the data of subtask 1.1 and 1.2 (\textit{CSD} in Figure \ref{feature_addition}),  reversing the sentences (\textit{RS}), generating additional data (\textit{GD}) and the pre-processing techniques from Section \ref{preprocessing}. Moreover, as in most machine learning problems, appropriately tuning the model hyperparameters also has a significant impact on the final score.

\subsection{Final results}

After presenting and analyzing the impact of each system feature separately, we show the overall results in this section. The final results on the official test set are presented on Table \ref{final_results}, ranking 1st in Subtasks 1.1, 1.2 and Subtask 2.C (joint result of classification and extraction) and 2nd for 2.E (relation extraction only). Furthermore, Table \ref{class-results} shows the differences in performance between relation types for Subtask 1.1.

\begin{table}[t!]
\centering
\begin{tabular}{|c|c|c|c|}
\hline
\textbf{Subtask}     & \textbf{P} & \textbf{R} & \textbf{F\tiny{1}}   \\\hline\hline
1.1         & 79.2  & 84.4  & 81.7 \\
1.2         &  93.3 & 87.7  & 90.4 \\
2.E    &  40.9 & 55.3  & 48.8 \\
2.C &  41.9 & 60.0  & 49.3\\\hline
\end{tabular}
\caption{Precision (P), recall (R) and $F_1$-score (F{\tiny1}) in \% on the test set by Subtask}
\label{final_results}
\end{table}

\begin{table}[t!]
\centering
\begin{tabular}{|c|c|c|c|}
\hline
\textbf{Relation type} & \textbf{P} & \textbf{R} & \textbf{F\tiny{1}} \\\hline\hline
COMPARE                & 100.00        & 95.24      & 97.56       \\
MODEL-FEATURE          & 71.01      & 74.24      & 72.59       \\
PART-WHOLE             & 78.87      & 80.00      & 79.43       \\
RESULT                 & 87.50      & 70.00      & 77.78       \\
TOPIC                  & 50.00         & 100.00        & 66.67      \\
USAGE                  & 87.86      & 86.86      & 87.36  \\\hline
Micro-averaged total   & 82.82      & 82.82      & 82.82       \\
Macro-averaged total   & 79.21      & 84.39      & 81.72      \\\hline
\end{tabular}
\caption{Detailed results (Precision (P), recall (R) and $F_1$-score (F{\tiny1})) in \% for each relation type on the test set for Subtask 1.1}
\label{class-results}
\end{table}

\section{Conclusion}

In this article we presented the winning system of SemEval 2018 Task 7 for relation classification, which also achieved the 2nd place for the relation extraction scenario. Our system, based on an ensemble of CNNs and RNNs, ranked first on 3 out of the 4 Subtasks (relation classification on clean and noisy data, and relation extraction and classification on clean data combined). We have tested various approaches to improve the system such as generating more additional training samples and experimenting with different order strategies for asymmetrical relation types. We demonstrated the effectiveness of preprocessing the samples by taking into account their length, marking the entities with explicit tags, defining an adequate surrogate optimization objective and combining effectively the outputs of several different models.


\clearpage

\bibliography{semeval2018}
\bibliographystyle{acl_natbib}

\end{document}